\documentclass{article}

\usepackage{PRIMEarxiv}
\usepackage{amsmath}
\usepackage{natbib}
\usepackage[utf8]{inputenc} 
\usepackage[T1]{fontenc}    
\usepackage{hyperref}       
\usepackage{url}            
\usepackage{booktabs}       
\usepackage{amsfonts}       
\usepackage{nicefrac}       
\usepackage{microtype}      
\usepackage{lipsum}
\usepackage{fancyhdr}       
\usepackage{graphicx}       
\graphicspath{{media/}}     

\usepackage{multirow}
\usepackage{makecell}
\usepackage{subfig}

\title{Tobler’s First Law in GeoAI: A Spatially Explicit Deep Learning Model for Terrain Feature Detection Under Weak Supervision
}

\author{
  Wenwen Li \\
  School of Geographical Sciences and Urban Planning \\
  Arizona State University \\
  Tempe, AZ, USA\\
  \texttt{wenwen@asu.edu} \\
   \And
Chia-Yu Hsu \\
  School of Geographical Sciences and Urban Planning \\
  Arizona State University \\
  Tempe, AZ, USA\\
   \And
     Maosheng Hu \\
 School of Geography and Information Engineering \\
 China University of Geosciences \\
  Wuhan, China\\
 }  

\begin{document}
\maketitle

\textbf{Abstract}: Recent interest in geospatial artificial intelligence
(GeoAI) has fostered a wide range of applications using artificial
intelligence (AI), especially deep learning for geospatial problem
solving. However, major challenges such as a lack of training data and
ignorance of spatial principles and spatial effects in AI model design
remain, significantly hindering the in-depth integration of AI with
geospatial research. This paper reports our work in developing a
cutting-edge deep learning model that enables object detection,
especially natural features, in a weakly supervised manner. Our work has
made three innovative contributions: first, we present a novel method of
object detection using only weak labels. This is achieved by developing
a spatially explicit model according to Tobler's first law of geography
to enable weakly supervised object detection. Second, we integrate the
idea of an attention map into the deep learning based object detection
pipeline, as well as develop a multistage training strategy to further
boost detection performance. Third, we have successfully applied this
model for the automated detection of Mars impact craters, the inspection
of which often involved tremendous manual work prior to our solution.
Our model is generalizable for detecting both natural and manmade
features on the surface of the Earth and other planets. This research
has made a major contribution to the enrichment of the theoretical and
methodological body of knowledge of GeoAI.


\section{Introduction}\label{introduction}

Machine learning represents an exciting new research area in artificial
intelligence (AI) that incorporates machine intelligence and data-driven
approaches for geospatial problem solving. Rapid advances in AI methods,
the proliferation of spatial big data, and the increasing availability
of computing power are transforming the way we conduct geospatial
research and prompting new discoveries. Geospatial artificial
intelligence (GeoAI) has emerged as a new research area that tackles
data- and computation-intensive problems leveraging AI and geospatial
big data \cite{li2020geoai}. One key topic in GeoAI research is spatial object
detection, the task to distinguish features, either natural or manmade,
in remote sensing and other forms of images. The derivation of such
information will greatly enrich existing gazetteers of both named and
unnamed features, thereby advancing our spatial knowledge about the
Earth and other planets \citep{hill1999gazetteer, goodchild2008gazetteer, zhu2020gcn}. Feature detection is also playing an important role in
urban planning \cite{kamusoko2017urban}, environmental management \cite{barrett2013satellite}, search and rescue operations \cite{bejiga2017cnn} as well as
inspection of the living conditions of refugee camps \cite{tomaszewski2016zaatari}.

Modern terrain analysis can also greatly benefit from advances in GeoAI.
Traditional approaches often involve the use of object-based image
analysis (OBIA) and some shallow machine learning techniques, such as
support vector machine or random forest, to find, segment, and classify
objects of interest in an image scene. While proved successful with
different detection problems \cite{jasiewicz2013geomorphons, micheletti2014landslide, arundel2018effect}, these approaches inevitably require
manual work; for instance, manual determination of attributes or terrain
factors that are important for distinguishing landform features of
different types. In OBIA, data processing parameters such as scale
factor, which controls the segmentation granularity, and the strategies
for merging segmented super-pixels (a cluster of pixels with similar
values), also need to be manually or semi-automatically selected. The
breakthroughs in machine learning, especially the rapid development of
deep learning techniques, have offered ample opportunity to
revolutionize the terrain analysis paradigm toward operating in a more
intelligent and automated manner \cite{li2017crater}.

One significant challenge in applying AI and deep learning for natural
feature detection, however, is the lack of proper training data. This is
due to the high cost and level of expertise required for collecting
training data of good quality. A survey shows that among the available
geospatial benchmark databases  \cite{yang2010bow, cheng2017scene, xia2017aid, zhou2018patternnet}, only four contain
terrain features of limited types. Moreover, existing datasets are
primarily used for scene classification instead of object detection.
They contain only labels of object types in each image; the extent of
each object is not delineated. A second major challenge is that many
deep learning applications using geospatial data are simply an
importation of methods from computer science to geography. There is very
limited work toward methodology innovation that incorporates spatial
principles and the unique characteristics of spatial data to supervise
the learning processes.

However, spatial theories, such as Tobler's first law of geography (TFL)
should be taken deeply into account in studies related to space and
place. TFL states, ``Everything is related to
everything else, but near things are more related than distant things.''
Essentially, it is a perfect, informal description of spatial
autocorrelation, the intrinsic relationships between geographical
entities. In fact, TFL has been guiding the design of many spatial
methods, such as local indicators of spatial autocorrelation (LISA;
\citep{anselin1995local, anselin2014metadata}) and geographically weighted regression \cite{fotheringham2003gwr}. To further exert the value of GeoAI and
deep learning in geospatial research in general and terrain analysis in
particular, this new machine learning--based paradigm needs to be
revamped to deeply integrate spatial theory and laws, such as TFL.

To address the above challenges, we developed a novel deep learning
model to enable weakly supervised object detection (WSOD). A WSOD task
aims to achieve object detection with only weak labels. Compared to
strong supervision, which requires object-level annotation (object class
and bounding box within each training image), weakly supervised learning
only requires image-level annotation (object classes and/or an object
count within a training image). Previous studies have shown that marking
weak labels takes less than 5\% of the time needed for marking strong
labels \citep{gao2018cwsl}. However, predicting object location without
explicitly providing this information will require the model to be
smarter than strongly supervised object detection models. Our proposed
solution accomplishes this goal by injecting Tobler's first law of
geography into the deep learning model design to achieve good detection
performance with a low labeling cost. The novel proposed strategy which
explicitly models the spatial relations, particularly spatial
autocorrelation allows us to tackle both problems (object detection and
lack of training data) within one solution framework. In addition, we
also integrated a machine vision strategy with an attention map and
devised comprehensive training strategies to further boost the WSOD
performance. Experiments were conducted using a large Mars crater data
set and a natural feature dataset on Earth, and the results showed that
our proposed approach achieved state-of-the-art performance in WSOD.

The remainder of this paper is organized as follows: Section 2 reviews
the current WSOD in both computer vision and the geography community.
Section 3 provides a formalized definition of the problem. Section 4
introduces how spatial explicitness is achieved in the model design.
Section 5 describes our methodology in detail. Section 6 describes the
experimental setting and results, and Section 7 concludes the paper and
discusses future research directions.

\section{Literature Review}\label{literature-review}

Methods for image analysis and natural-feature detection can be
classified into two broad categories: knowledge-driven and data-driven.
Below we provide an overview of the two approaches and discuss in detail
the new deep learning technique and its usage in remote sensing.

\subsection{Knowledge-driven terrain and image
analysis}\label{knowledge-driven-terrain-and-image-analysis}

A knowledge-driven approach leverages prior knowledge or existing
patterns to design an algorithm that performs various tasks. An
advantage of such approaches is that they are based on a top-down,
expertise-driven design pattern, so the data processing workflow, i.e.,
how it works and why it works, is well reasoned and often transparent
and is therefore easy to explain. Specifically, current terrain analysis
leverages different techniques to identify natural features in image
data. For instance, thresholding is a commonly used method in analyzing
terrain. It can be considered a binary classification process, in which
a threshold is applied to terrain parameters, such as elevation, slope,
curvature, or the numerical difference between them, to help classify a
pixel into an object or a non-object \citep{blaschke2010object}. Stream or drainage
network analysis is another popular approach for extracting terrain
features. By simulating how water accumulates at lower elevations in a
watershed, a stream network can be created. Using the same approach, the
valleyline features can be extracted from a digital elevation model
(DEM), and so can ridgelines. The only difference is that for ridgeline
extraction the elevation values need to be reversed first to form a
pseudo-watershed \cite{lindsay2015lidar}. In recent years,
spatial-contextual-based analysis, which formulates the spatial
relationships between nearby pixels and the contextual background of a
spatial object, has become an important form of expert knowledge that
further improves object-based image analysis (OBIA) for feature
segmentation. Zhou et al. \citep{zhou2019spatio} developed a visual descriptor that
captures the spatial-contextual pattern through a probabilistic model to
quantify the patterns of linear terrain features on a DEM, and the
approach has proven to be effective in identifying ridges and valleys in
mountainous regions.

Although popular, these knowledge-driven approaches have been facing
great challenges in terrain analysis, especially when dealing with big
and high-resolution data. First, terrain features are complex, and they
often possess inter-category similarity and intra-category
heterogeneity. In addition, the same terrain feature may demonstrate
different characteristics in different landscapes, making it very
difficult to extract a common pattern to support its identification.
Second, super-high and high-resolution terrain data, which have become
increasingly available, also present major challenges to existing
methods. Although these data provide more details about the terrain, the
high-resolution data also capture more ``noise,'' i.e., local terrain
spikes due to vegetation or a rocky surface. Existing methods, which are
designed to process ``smoothed'' data, have shown limitations in
handling data with uncertainties. Third, the knowledge-driven approaches
often need accumulations of knowledge over a long time. They therefore
require human intervention and are difficult to fully automate. To
alleviate the aforementioned issues, data-driven approaches, such as AI
and machine learning, have attracted researchers' close attention and
become the fourth paradigm in scientific research \citep{gahegan2020gis}.

\subsection{Data-driven object detection leveraging Weakly
Supervised Object detection (WSOD) in computer
vision}

In recent years, deep learning has emerged as the cutting-edge
machine-learning framework for object detection \citep{li2022geoai}. It is known as the outstanding ability to learn context features of real-world objects from
large quantities of labeled data. Typically, detectors are trained under
strong supervision with object-level annotations. However, it is very
time consuming for drawing bounding boxes (BBOX) manually. On the other
hand, object detectors can also be trained under weak supervision using
only image-level labels, the collection cost of which is substantially
lower than obtaining object-level labels. The most popular pipeline for
WSOD has three steps: feature extraction, proposal generation and
proposal classification. Note that in the context of object detection, a
``proposal'' refers to a candidate bounding-box (BBOX) that covers the
object in an image.

Based on this pipeline, different works have been introduced to improve
model performance in each step. Bilen and Vedaldi \cite{bilen2016weakly} introduce one of phenomenal works -\/- WSDDN (Weakly Supervised Deep Detection
Networks) that use a pre-trained CNN (Convolutional Neural Network) for
WSOD. This model assumes that a pre-trained CNN generates meaningful
representations of the data which even contains location information of
the entire and part of an object \cite{zhou2014deep}. Based on this
assumption, WSDDN is designed with a two-stream structure to explicitly
reason about image regions. However, the limitation of this model is its
tendency to generate BBOX that contain only the part instead of the
entire object. Tang et al. \citep{tang2017multiple} developed a new model named OICR
(Online Instance Classifier Refinement) to leverage multi-stage learning
to continue refining the BBOX to eliminate the issue of partial-labeling
caused by WSDDN. Each stage is a classifier trained by the outcome of
the previous stage. As a result, the latter stages are trained not only
with the most critical parts but also the parts overlapping with them.
Besides the iterative refinement of BBOX, the initial selection of
candidate proposals (BBOX) is also important. C-MIDN (Gao et al. 2019)
is a cutting-edge model that employs a parallel WSDDN structure and a
segmentation map for selecting better proposals at the first stage of
the optimization pipeline.

Almost all of the above works use traditional proposal generation
methods, such as selective search \citep{uijlings2013selective} or edge boxes
\citep{zitnick2014edge} to create a large number of BBOX and then
refine them until an optimal BBOX is found. There are much fewer works
focusing on making improvement on region proposal network, which aims to
generate more accurate candidate BBOX in the first place \citep{hsu2021learning}. High quality proposals, however, have been proved to have a great
influence on the performance of object detection tasks \citep{hosang2015proposals}. One pioneer research belonging to this second kind is the work by
Tang et al. \citep{tang2018weakly}, which propose a multi-stage region proposal network
where each stage filters out part of the original proposals based on
different strategies. However, the model design is quite complex
involving the cascading of two WSOD models trained separately, which
inevitably increases training time and is difficult to be reproduced.
Our work also focuses on making improvement on proposal generation.
Unlike the work by Tang et al. \citep{tang2018weakly}, our proposed deep neural network
generates high quality and precise proposals directly instead of a
selection from randomly generated proposals.

\subsection{Applications of object detection in remote
sensing}\label{applications-of-object-detection-in-remote-sensing}

Object detection can find a wide range of applications across urban and
environmental science domains. Many recent studies work on tackling the
challenges of object detection from remote sensing imagery, for example,
Li and Hsu \cite{li2020terrain} extends Faster R-CNN \citep{ren2015faster} to enable
natural feature identification from remote sensing imagery. The authors
evaluated performance of multiple deep CNN models and found that the
very complex and deep CNN models do not always yield the best detection
accuracy. Instead, the CNN model should be carefully designed according
to the characteristics of training data and complexity of the objects
and background scene. Other issues and studies like rotation-sensitive
detection \citep{ding2018roi, liao2018rotation, zhang2019double, yu2015rotation, li2017rotation, cheng2018rotation, cheng2016vhr},
proposal quality \citep{long2017cnn, zhong2018multiclass, xu2017deformable}
and real-time object detection \citep{liu2015vehicle,tang2017multiple} 
are also developed.

In addition, weakly supervised learning and target detection from remote
sensing image has been exploited by geospatial researchers. Han et al.
\citep{han2014weakly} use multi-scale sliding window to generate proposals from images and then leverages a Bayesian classification network to classify
proposals based on hand-crafted features, including low-level local
feature extracted from segmentation, statistics of these low-level
features in an image patch (mid-level feature) and high-level features
from deep Boltzmann machine. Zhang et al. \citep{zhang2014weakly} achieves weakly
supervised learning by separating images into positive (with targets)
and negative (without targets) samples without BBOX information. The
positive images are initially obtained using saliency-based
self-adaptive segmentation and then refined by negative mining to remove
positive images without targets. This method is further improved by
carefully selecting negative images which are informative, tend to be
misclassified, diverse, and non-redundant \cite{zhou2016weakly}. A
pre-trained CNN is used to extract discriminative features. These works
yield interesting results, but are limited in single-class detection.
Hence, their applicability in dealing with a large dataset and complex
object detection tasks still need to be verified. In comparison, our
work relies on a deep learning framework, the entire learning process --
from feature extraction, to proposal generation and classification are
all automatically done. Next section presents the formal problem
statement.

\section{Problem Formulation}\label{problem-formulation}

In scenarios of strongly supervised object detection, the problem is to
predict object-level labels, including object class
(\(O_{i,\ \ Class}\)) and object bounding box (BBOX; \(O_{i,BBOX}\)),
given the same information in the training data. In our study, we
enabled a WSOD by replacing the BBOX information with simply a count of
the objects in an image. Assuming \(i\ (i \in \lbrack 0,m\rbrack)\) is
the index of an object in an image, and \(j\) is the index of an image
in the training samples, our WSOD problem can be formally represented as
finding the mapping \(f\) such that

\begin{equation}
f: \langle \text{Image}_{j,\{\text{Class}\}},\ m \rangle \rightarrow \langle O_{ij,\text{Class}},\ O_{ij,\text{BBOX}} \rangle
\end{equation}

where \(m\) is the object count in an \(Image_{j}\), and \(\{ Class\}\)
is all the object classes in the same image; \(O_{ij}\) is the index of
an object in \(Image_{j}\). As previous studies have shown, in the
labeling process, the time for counting is substantially less than
drawing the actual BBOX. Our research will therefore directly tackle the
issue of high label cost in collecting training data, which is essential
for supervised learning to achieve satisfying predictive performance. We
accomplished this by explicitly incorporating spatial theory and
principles (i.e., spatial autocorrelation) in the model design, which is a brand-new attempt in GeoAI
and deep learning.

\section{Spatial Explicitness in Deep Learning
Models}\label{spatial-explicitness-in-deep-learning-models}

A key issue that a WSOD network must solve is to identify the object
location without the provision of this information in the training data.
This location information can be represented as ``critical points'' that
stay on or near an object such that a candidate BBOX can be drawn around
these points for proposal generation. Here, we propose leveraging
temporal data classification, such as with a recurrent neural network
(RNN), to perform object detection in a weakly supervised manner.
Current RNNs, such as long short-term memory (LSTM) have shown
outstanding performance in classifying one-dimensional (1D) sequence
data, such as a speech segment. However, it requires per-frame labels in
its learning process. To reduce the required labeling information, we
further extended the LSTM by giving it a new objective function, namely
connectionist temporal classification (CTC), which learns to identify
the optimal segmentation location to separate different words in a
speech sequence with only per-sequence labels, meaning using, for
example, ``Hello world'' for a speech segment rather than labels for
individual letters, ``H-e-l-l-o-w-o-r-l-d,'' on each speech frame. Note
that each data slice that forms a temporal/sequential data could be
letters in a text document, phonemes in a speech segment or sequential
image data, i.e. time series remote sensing data or a video clip. These
data can all be fit into a temporal classification framework.

One key question in enabling the 2D WSOD using this 1D temporal
classification strategy is proper dimension reduction. Here, we argue
that spatial theory and principles (i.e., spatial continuity and spatial
autocorrelation) play a key role in the applicability of 1D temporal
classification for two-dimensional (2D) object detection. According to
Tobler's first law of geography \cite{tobler1970urban}, which states that nearby
things are more related to each other, we can perform a serialization on
the 2D data by applying row-prime or column-prime scan orders. After
serialization, spatial continuity in the main scanned direction will be
retained. Although the spatial continuity perpendicular to this
direction will be broken, but because LSTM can capture contextual
information in both immediate neighbors (short-term memory) and distant
neighbors (long-term memory), this broken spatial dependency can still
be ``memorized'' by the network.

\begin{figure}[ht]
  \centering
  \includegraphics [width=0.8\textwidth]{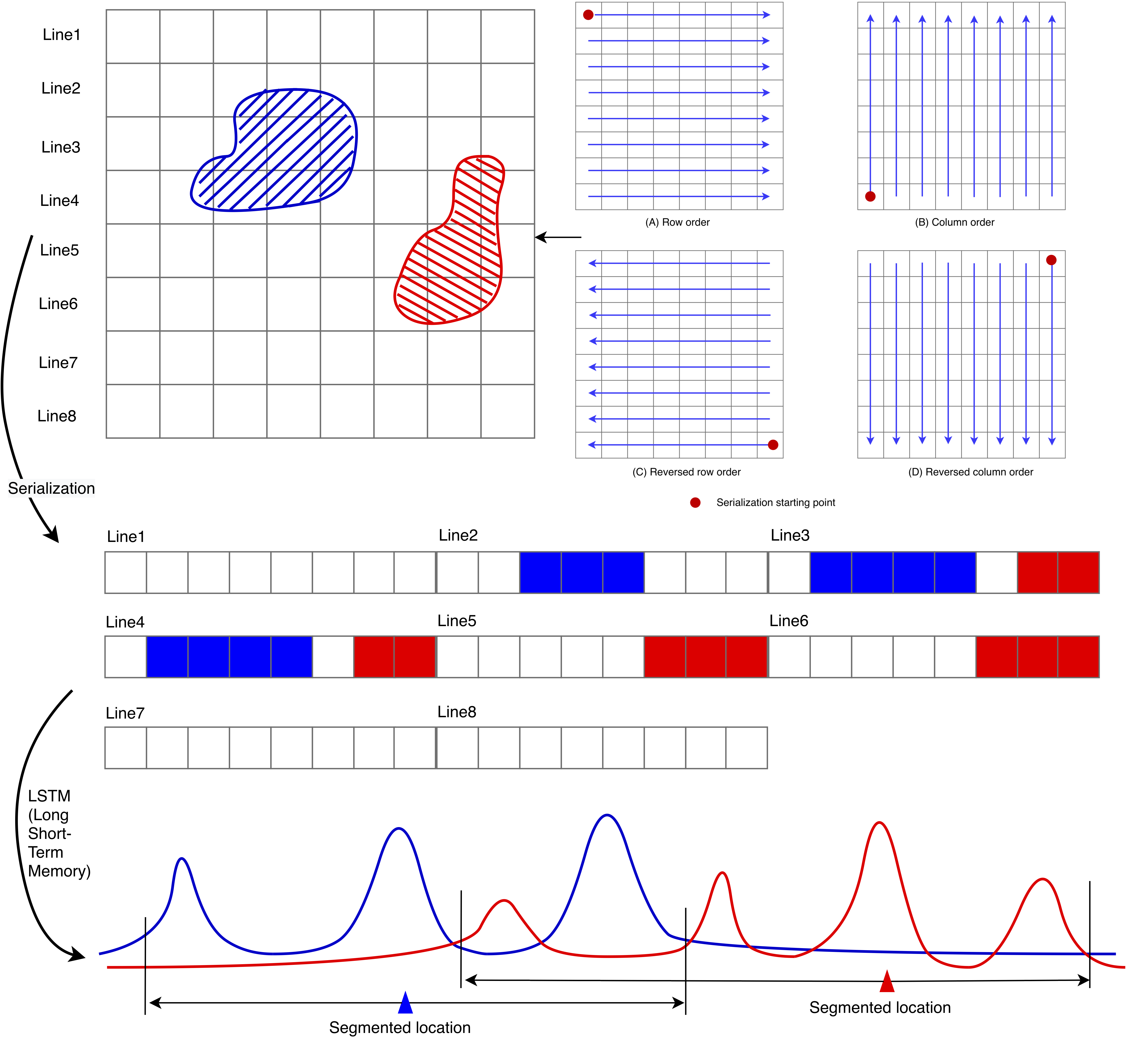}
  \caption{A visual example of a spatial theory enabled strategy for weakly supervised object detection.}
  \label{fig:figure1}
\end{figure}

Figure \ref{fig:figure1} illustrates the spatial theory enabled WSOD. Given a feature map, we proposed to apply scan orders at four different directions
(row-order, reserved row-order, column-order, and reversed column-order)
to serialize the feature map into a 1D feature sequence. The data were
then sent to the temporal classification model to identify the optimal
segmented locations for the objects. Our model differs from the
traditional LSTM-CTC framework in two ways: first, a CTC requires labels
of each word (both ``hello'' and ``world'') in a phrase (``hello
world''). Our model instead takes the segmentation as a binary problem
in which we treat objects of different types the same way; hence, there
is only a need to provide a total object count (2 for ``hello world'').
Our goal is to identify and separate the BBOX of all objects of interest
(the foreground) from the background scene. Then, a CTC training process
identifies the optimal mapping of predicted objects on the feature
sequence to the object count information. Second, the objective of the
CTC is to find where the words are located in each sentence and their
order of occurrences; hence, the segmentation results based on
probability estimation are always the location labels of each word. When
applying this mechanism for segmenting 2D objects with one spatial
dimension broken, the segmented location will tend to locate on the most
prominent spike (area with high visual attention) of the feature
sequence. We call this location (or a series of such locations) the
``critical point(s)'' of an object, and it will be used as the center
point for generating candidate proposals (BBOX) for object
classifications in the next stage. In our proposed model, the feature
maps are serialized in four different directions (as shown above)
because the object appearing in the feature maps may exert different
temporal patterns. When training four temporal classification models
(e.g., LSTM) with a shared CNN, behaviors of the LSTMs will be
influenced by each other, and the four networks will tend to converge on
the points falling on the same set of objects of interest, thereby
generating more accurate predictions about object locations. Different
from the commonly used selective search strategy, which needs to draw a
great number of proposals (for instance, 2,000) to exhaust the potential
locations and sizes of an object, our proposed approach can more
intelligently locate the suspected area where an object will appear, and
therefore substantially reduce the number of proposals while increasing
their quality. After the proposals are generated, they are sent to a
region-based classifier for proposal ranking, refinement, and
classification. Any region-based classifier under weak supervision could
be integrated with this proposal generation network for object
localization and prediction.

\section{Methodology}\label{methodology}

\subsection{Weakly Supervised Object Detection
Pipeline}\label{weakly-supervised-object-detection-pipeline}

\begin{figure}[ht]
  \centering
  \includegraphics [width=1\textwidth]{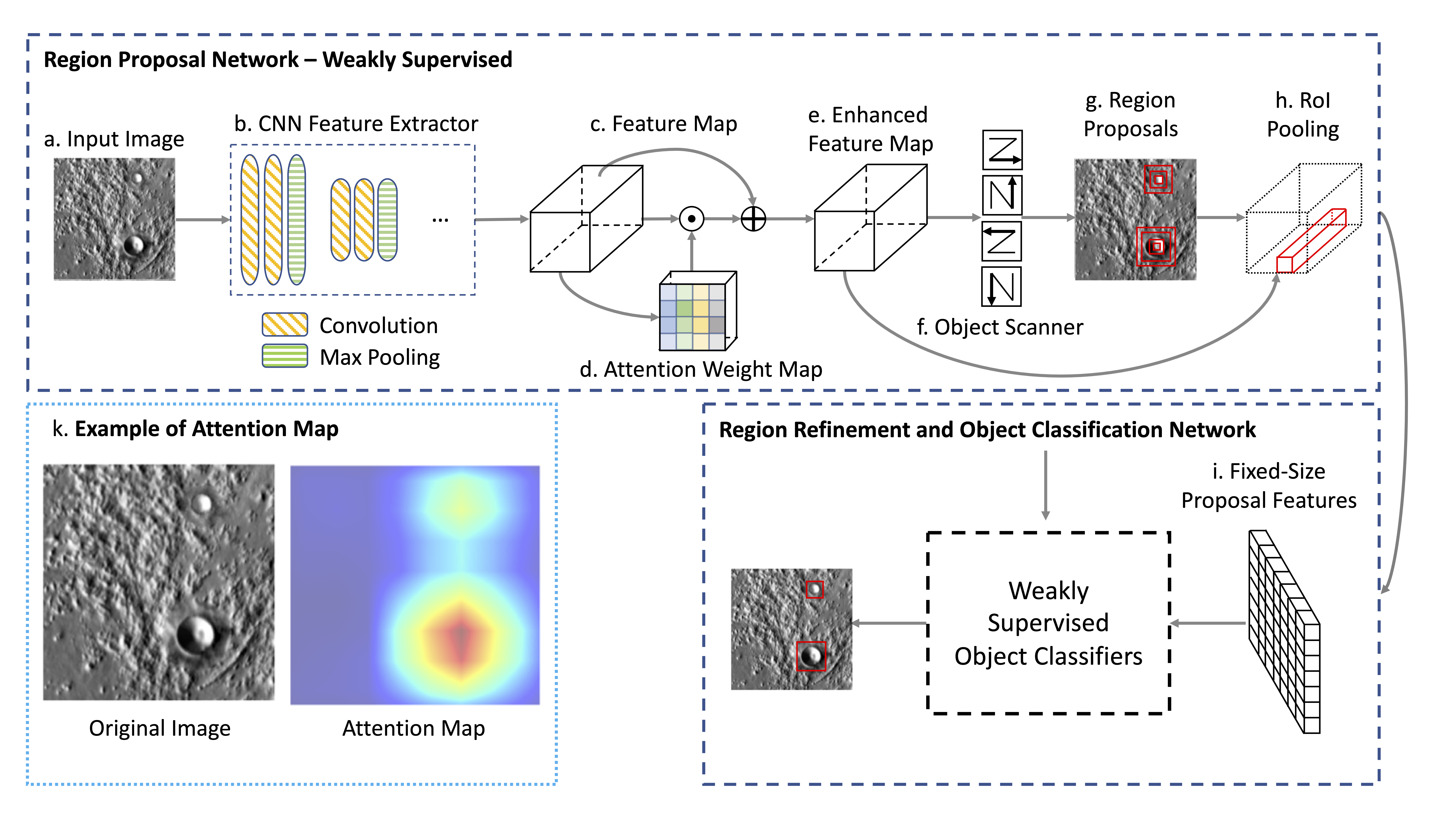}
  \caption{Proposed weakly supervised object detection (WSOD) pipeline. CNN = Convolutional neural network; RoI = region of interest.}
  \label{fig:figure2}
\end{figure}

Figure \ref{fig:figure2} illustrates the model structure and workflow of the proposed
WSOD network. This network achieves active learning in two stages. The
first stage is a region proposal network (RPN), which leverages the
aforementioned temporal classification to identify candidate object
proposals (upper pipeline). The second stage (lower pipeline) refines
these proposals and conducts object classification. Instead of using
strong instance-level supervision, both networks are supervised with
weak labels, including a total object count and image-level annotations
on object classes.

At the region proposal phase, the input image first goes through a CNN
module (Figure \ref{fig:figure2}b) for extracting the most prominent features that
distinguish different images and objects. The resultant product is
termed a feature map,
\(\mathbf{X} \in \mathbb{R}^{H \times W \times D}\) (H, W, and D are the
height, weight, and depth of the feature map, respectively). Before
sending the feature map\(\mathbf{\ X}\) for subsequent processing, we
augmented the feature map to highlight the most informative and relevant
subregions. This was achieved by an attention mechanism that simulates
how humans pay visual attention to different areas of an image once a
prompt is given. Following this mechanism, a spatially normalized
attention weight map \(\mathbf{A} \in \mathbb{R}^{H \times W}\) is
generated (Figure \ref{fig:figure2}d). The values in \(\mathbf{A}\) are normalized such
that the sum of all values in \(\mathbf{A}\) equals 1. Next, the
original feature map \(\mathbf{X}\) is multiplied (channel-wise) by the
weight map \(\mathbf{A}\) to create attention map
\(\mathbf{X}_{a} \in \mathbb{R}^{H \times W \times D}\). The enhanced
feature map \(\mathbf{X'}\) (Figure \ref{fig:figure2}e) is created by adding
\(\mathbf{X}_{a}\) and \(\mathbf{X}\) together.

Once the enhanced feature map is generated, it is sent to the core
module of the RPN---the object scanner (Figure \ref{fig:figure2}f), which contains four
parallelly running temporal classification modules empowered by LSTM
with CTC as the objective function. The four modules take serialized
feature maps by the proposed scan orders as inputs (as shown in Figure
1) and output the segmented locations, which we call critical points, on
or near the target objects. Then, multiple BBOX at different ratios and
shapes are generated around the critical points to serve as the
candidate region proposals (Figure \ref{fig:figure2}g) for future refinement and object
classification.

Next, the region of interest (RoI) pooling layer (Figure \ref{fig:figure2}h) will
extract and generate fixed-size feature maps of proposed regions (Figure
\ref{fig:figure2}i) from the enhanced feature map and send it to the object classifier
(Figure \ref{fig:figure2}, bottom pipeline). The common workflow of the object
classifier is to make predictions on the class of object within the
candidate proposals. Because the deep network is weakly supervised,
there is often a mechanism to rank and refine the proposals to increase
the detection accuracy. Many deep learning models are designed to
achieve this goal, such as the work by Bilen et al. \citep{bilen2016weakly}, Tang et al.
\citep{tang2017multiple}, and Gao et al. \citep{gao2018cwsl}, among others. However, almost all these
models target the improvement of the object classifier (Figure \ref{fig:figure2},
bottom). Conversely, our model focuses on improving the RPN (Figure \ref{fig:figure2},
top) by incorporating spatial theory and principles and then utilizing
existing models for object classification. It is worth noting that our
framework is designed flexibly enough so it can be easily integrated
into other models by replacing their RPN with our proposed network or by
integrating the object classifiers of other models into our detection
pipeline.

Taking the coupled multiple instance detection network (C-MIDN; \citep{gao2019cmidn} that is used as the object classifier in our task as an
example (Figure \ref{fig:figure2}, bottom), it partitions the object classification into
two phases: initial detection and detection refinement. In initial
detection, two proposal classifiers are applied; each contains two
branches, with one responsible for detection and the other for
classification. The first classifier will select top-scoring proposals
as the detection result. However, these proposals may not always be of
high quality (e.g., they may contain only partial objects). To address
this issue, a segmentation map is integrated for removing poor-quality
proposals, even if they are ranked highly but have little overlap with
the detected area of interest (the foreground in the segmentation map).
The second classifier will take the filtered proposals to perform
classification with the aim of finding better results.

After the initial detection, the proposals and classification results
are sent to a multistage refinement network that is also deep learning
based. Each stage is trained under the supervision of instance labels
obtained from the previous stage. To obtain instance labels for
supervision, given an image with class label c, the proposal \(j\) with
the highest score for class c will be used as the pseudo ground truth
BBOX. In addition to labeling \(j\), other proposals that have a high
spatial overlap with \(j\) will also be labeled as class \(c\).
Meanwhile, proposals that do not belong to class \(c\) will be labeled
as background. Using this refinement strategy, the detected BBOX and
object classification results can be further enhanced.

\subsection{Region Proposal Network
(RPN)}\label{region-proposal-network-rpn}

\begin{figure}[ht]
  \centering
  \includegraphics [width=1\textwidth]{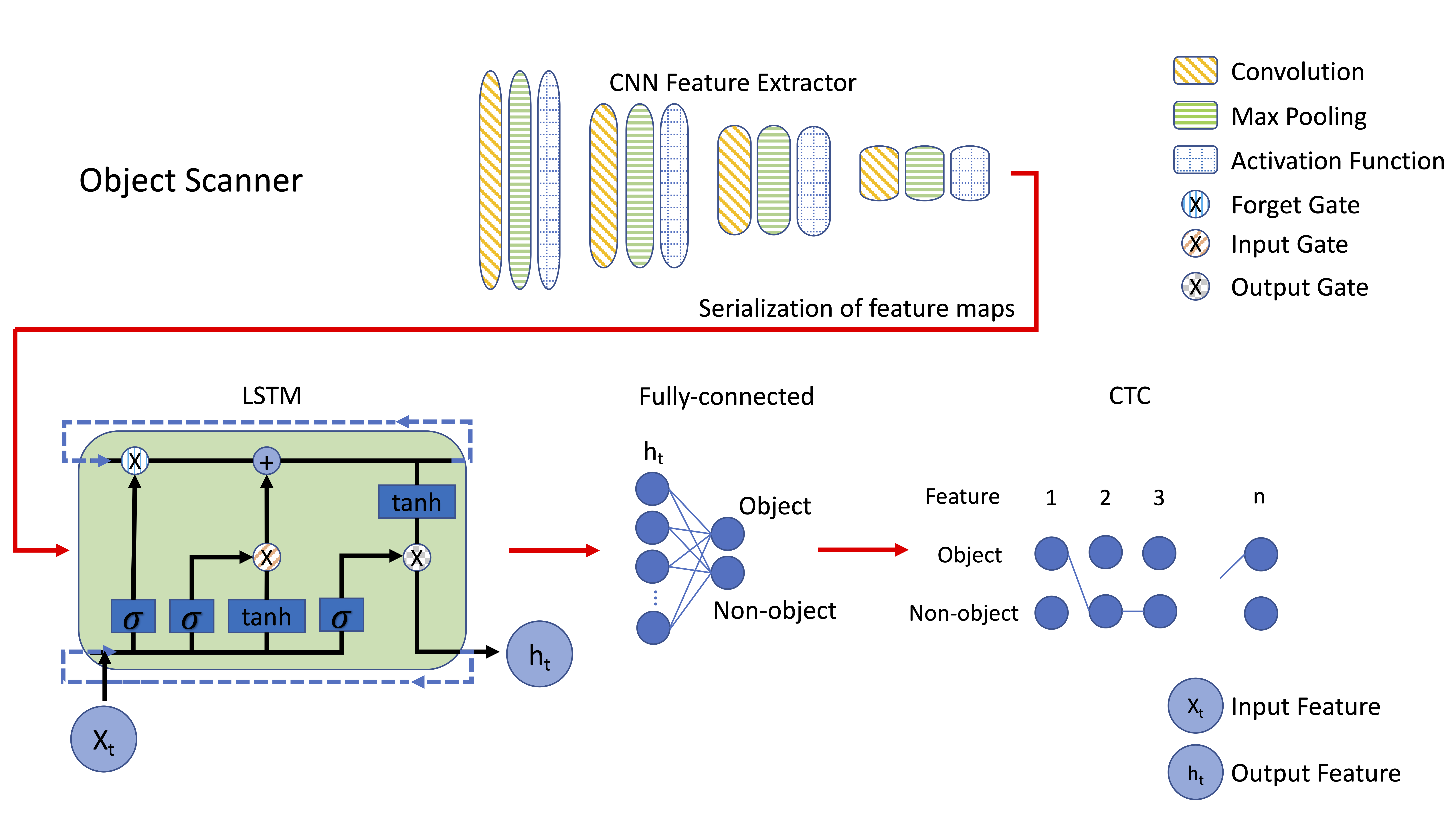}
  \caption{A spatially explicit design of the object scanner. CNN = Convoluted neural network, CTC = connectionist temporal classification, LSTM = long short-term memory}
  \label{fig:figure3}
\end{figure}

As introduced, the RPN consists of four object scanners (Figure \ref{fig:figure3})
integrated in parallel. Scanners are responsible for locating the most
discriminative part of objects and each consists of four core
components: a joint CNN, an LSTM, a fully connected layer, and a CTC
layer. For each image, the CNN extracts low-, mid-, and high-level
features through consecutive convolution operations and generates
so-called feature maps that contain latent features of candidate objects
in the image. Next, serialization is applied on the feature maps,
transferring 2D maps into 1D sequences \(\mathbf{x}\), where
\(\mathbf{x} = \left( x_{1},\ x_{2},\ \ldots,\ x_{T} \right)\),
representing the feature sequence after serialization. There are four
different serialization orders (row-prime, column-prime, reversed
row-prime, and reversed column-prime), and each scanner uses one of
them. By this transformation, the 2D spatial relations between objects
are converted into four 1D sequential relations. These 1D sequences are
fed into LSTMs, networks that persist historical information to leverage
global spatial context for discriminative object detection. Different
from classic object detection models, which are based on analysis of the
2D images and their transformed feature maps after layers of
convolution, our proposed deep learning model follows a very different
path, which is to leverage temporal classification performed on 1D
sequential data for the object detection task.

The flow of information in LSTM is regulated by gates, which is a
sigmoid operation followed by a pointwise multiplication operation. The
gate control signals are a concatenation of current input and the output
from the previous feature vector. There are three gates in each LSTM, an
input gate, an output gate, and a forget gate. A forget gate controls
the amount of information that will flow into the current state from the
previous state, an input gate controls the percentage of the current
input that will be added to the output of the previous state, and an
output gate controls the output amount from the current state. These
three gates make LSTM capable of learning long-term spatial dependencies
that have been broken by the serialization and, therefore, can be
leveraged to predict the objects' most discriminative parts.

More formally, the object recognition process can be described as
follows: Given a feature sequence
\(\mathbf{x} = \left( x_{1},\ x_{2},\ \ldots,\ x_{T} \right)\), our
target is to find a label sequence
\(\mathbf{L}^{\mathbf{*}} = \left( l_{1},\ l_{2},\ \ldots,\ l_{U} \right)\)
that annotates \(\mathbf{x}\). Each \(l\) comes from a finite alphabet
set (e.g., a to z for speech classification), and the length of the
label sequence \(U\) will be equal to or shorter than the original
sequence length \(T\) because the sequence may also contain non-class
elements. When utilizing the 1D sequence classification for 2D WSOD,
each \(x\) is an \(m\)-dimensional vector serialized from the 2D feature
map with \(m\) channels;
\(l_{i} \in \left\{ 0,1 \right\}\ (\ i \in \lbrack 1,U\rbrack),\ \)
refers to the labels location for objects of interest. In our
count-supervised learning context, all objects are considered as a
single class (foreground), and the non-object locations are annotated as
0, referring to the background. Our objective has then become to find
the optimal placement of 1s in the feature sequence \(\mathbf{x}\) such
that \(\sum_{i = 1}^{U}{l_{i} =}n\), where \(n\) is the object count in
an image.

More formally, let \(\mathbf{y} \in {\mathbb{(R}^{2})}^{T}\) be the
output of the LSTM network, and \(\mathbf{y}\) is denoted by
\(y_{t}^{k}\), where \(k \in \left\{ 0,1 \right\}.\) In addition,
\(\ y_{t}^{1}\) denotes the probability of \(y_{t}\) being a critical
point on an object of interest, and \(y_{t}^{0}\) means at time \(t\);
\(y\) is not activated and is, therefore, a background element. A
typical CTC assumes that the network outputs \(\mathbf{y}\) at different
times \(t\) are conditionally independent; hence, the possibility of the
original sequence \(\mathbf{x}\) to contain a label sequence
\(\mathbf{\pi}\) can be expressed as

\begin{equation}
p\left( \boldsymbol{\pi} \middle| \mathbf{x} \right) = \prod_{t = 1}^{T} y_{t}^{k},\quad \forall \mathbf{k} \in \{0,1\},\quad \boldsymbol{\pi} \in L^{T}
\tag{2}
\end{equation}

where \(L^{T}\) is the set of all possible per-frame label sequences
that yield the final labels \(L^{*}\). To avoid the occurrence of
repeated labels for the same object, a many-to-one mapping of
\(\mathcal{B}\) from \(L^{T}\) to \(L^{*}\) is defined to remove the
continuously repeated labels. Finally, \(\mathcal{B}^{- 1}\) is
leveraged to identify the probability of a given label as the sum of all
possible \(\mathbf{\pi}\) that yields the same \(L^{*}\):

\begin{equation}
p\left( L^{*} \middle| \mathbf{x} \right) = \sum_{\boldsymbol{\pi} \in \mathcal{B}^{-1}(L^{*})} p\left( \boldsymbol{\pi} \middle| \mathbf{x} \right)
\tag{3}
\end{equation}

The objective function can therefore be written as:

\begin{equation}
O_{\mathrm{RPN}} = -\ln\left( p\left( L^{*} \middle| \mathbf{x} \right) \right)
\tag{4}
\end{equation}

Different from the objective of the original CTC, instead of predicting
the occurrence of different objects, our new model considers all objects
to be the same type (foreground). Hence, each input element in the
feature sequence will be predicted with the probability of being 1
(foreground) or 0 (background). The final prediction is to select
\(\mathbf{\pi}^{\mathbf{*}}\) which maximizes the probability of the
mapping from \(\mathbf{x}\) to \(L^{*}\), where

\begin{equation}
\boldsymbol{\pi}^{*} = \arg\max_{\boldsymbol{\pi}} p\left( \boldsymbol{\pi} \middle| \mathbf{x} \right)
\tag{5}
\end{equation}

The units in \(\mathbf{\pi}^{\mathbf{*}}\) that are classified as 1 are
the critical points predicted to be objects of interest. These locations
will serve as the center for generating multiple candidate proposals in
different sizes, which are then used to perform proposal refinement and
object classification.

\subsection{Proposal Refinement and Object Classification
Network}\label{proposal-refinement-and-object-classification-network}

The object classifier is trained to sort proposals from the RPN into
different classes. Since there are no object-level annotations, the
candidate proposals generated are a good guess and will need to be
further refined for more accurate localization. Here, we adopt the
object classifier in the C-MIDN model to perform the proposal refinement
and classification. There are three components in the classifier:
proposal classification, initial selection, and optimization.

\textbf{Proposal Classification:} Given region proposals \(\mathbf{R}\)
in the images, an RoI pooling layer extracts the corresponding features
and generates fixed-size proposal features. To generate image-level
predictions, the proposals from each image are branched into two data
streams, each generating a score matrix at the dimension of
\(|C|\mathbf{\times |}R\mathbf{|}\)\textbf{,} where \(C\) is the set for
classes and \(R\) is the set for proposals. These two score matrices
latter pass through a normalization process to generate two probability
matrices, which will be continuously refined during the training phase
to achieve different objectives. The one normalized along the row
(dimension of object class) is a matrix that indicates the location
probability distribution of a given class across all regions. Another
matrix, normalized along the column (dimension of different proposals),
stands for the probability distribution that a given region contains an
object across all classes. The difference between these two matrices is
that an object \(c\) could have a higher probability of being located in
a certain region \(r\) than in other regions, but this region may
contain an object other than \(c\). For instance, a crater may be more
likely to appear in a region where there are circular features, but
these features may come from other classes, such as islands or lakes.
Hence, a high score will be found at the class \emph{c} row and the
region \emph{r} column in the former matrix but will have a low score in
the same cell in the latter matrix.

The multiplication of corresponding cell values from two matrices can
actually imply the true probability of a region containing a given
class. A final matrix \(\mathbf{p}_{cr}\) for generating image-level
prediction is computed by element-wise multiplication of the two
matrices. Then, we can obtain a prediction score for a given class \(c\)
by the summation of the probability of this class over all proposals:

\begin{equation}
\varphi_{c} = \sum_{r = 1}^{|\mathbf{R}|} \mathbf{p}_{cr}
\tag{6}
\end{equation}

and because we have a prediction score for each class \(\varphi_{c}\)
and image-level labels
\(\{{\mathbf{y}_{c}\mathbf{\ |\ y}}_{c} \in (0,1)\}\), the objective
function is simply a multi-class cross entropy function:

\begin{equation}
O_{\mathrm{classifier}} = - \sum_{c = 1}^{|\mathbf{C}|} \mathbf{y}_{c} \log \varphi_{c}
\tag{7}
\end{equation}

\textbf{Initial Selection:} The proposal classification component gives
each proposal a class label and its related score; however, the highest
score is often localized at the most discriminative part of an object
instead of the entire object. Two proposal classification components can
be coupled. The top-scoring proposals of the first proposal classifier
are removed from serving as the input of the second one, pushing the
second classifier to search for other better proposals instead of
localizing the same proposal again. Furthermore, in case the first
classifier already locates the full-context proposal, a segmentation map
of a given class in the image is employed to improve the robustness of
the proposal removal process. If the overlap between the highlighted
areas in the segmentation map and the top-scoring proposal is too small,
it is more likely that there exists a better proposal. Suppose the set
of pixels in the segmentation map for class \(c\) is \(M_{c}\), and the
set of pixels in the top-scoring proposal from the first proposal
classification component is \(N_{c}\), then the overlap \(r_{c}\) is
defined as

\begin{equation}
r_{c} = \frac{\left| M_{c} \cap N_{c} \right|}{\left| M_{c} \right|}
\tag{8}
\end{equation}

If \(r_{c}\) is smaller than a given threshold, the corresponding
proposal will be removed from the input of the second proposal
classifier; otherwise, it will be retained. In addition, the other
candidate proposals that have an intersection over union (IoU) larger
than a given threshold with the top-scoring proposals will also be
removed. After the removal process, the two classifiers will be trained
with the same objective function. The only difference is that the input
proposals are different. The final predicted proposal will be determined
through a selection process of the outputs of the two classifiers. If
the top-ranking proposals have little overlap, both proposals will be
retained. If the top-ranking proposals have a high overlap (over 50\%
IoU), the proposal from the second classifier will be selected because
the proposal tends to contain the entire object rather than its part.

\textbf{Optimization:} After the initial detection, the proposals and
classification results are sent to a refinement network (Figure 2m) that
is also deep learning based for further refinement of the detection
result. Each stage is trained under the supervision of instance labels
obtained from the previous stage. At iteration \(k\ (k > 0)\), the
instance classifier generates a proposal score matrix
\(\mathbf{p \in}\mathbb{R}^{(|C| + 1)\mathbf{\times |R|}}\), where \(C\)
is the set for classes and \(R\) is the set for proposals. The extra row
in the class dimension refers to non-object background. To obtain
instance labels for supervision, given an image with class label c, the
proposal \(j\) with the highest score for class c will be used as the
pseudo ground truth BBOX. In addition to labeling \(j\), other proposals
that have a high spatial overlap with \(j\) will also be labeled as
class \(c\). Meanwhile, proposals that do not belong to class \(c\) will
be labeled as background. Mathematically, the objective of this
refinement process can be expressed as:

\begin{equation}
O_{\mathrm{refinement}} = - \frac{1}{|\mathbf{R}|} \sum_{r = 1}^{|\mathbf{R}|} \sum_{c = 1}^{\mathbf{C} + 1} \mathbf{y}_{r} \log \mathbf{p}_{c,r}
\tag{9}
\end{equation}

, where \(c\) is class index and \(c \in C\), \(r\) is proposal index
and \(c \in R\), \(\mathbf{y}_{r}\) is the class label for proposal
\(r\), \(\mathbf{p}_{c,r}\) is the probability for a proposal \(r\) to
contain an object of class \(c\). Using this refinement strategy, the
detected BBOX and object classification results can be further enhanced.

The overall objective function of this WSOD model (\(O_{WSOD})\) has
become the addition of objectives in the RPN phase (\(O_{RPN}\) in Eq.
4), the object classification phase (\(O_{classifier}\) in Eq. 7) and
the refinement phase (\(O_{refinement}\) in Eq. 9). Namely,

\begin{equation}
O_{\mathrm{WSOD}} = O_{\mathrm{RPN}} + O_{\mathrm{classifier}} + O_{\mathrm{refinement}}
\tag{10}
\end{equation}

The solution of this problem will toward minimizing \(O_{WSOD}\). Hence
\(O_{WSOD}\) can also be used as the loss function.

\section{Results and Discussions}\label{results-and-discussions}

\subsection{Data Preparation---Mars Crater Data
Set}\label{data-preparationmars-crater-data-set}

To assess the performance of our proposed model, we created a Mars
crater data set that contains 10,000 image scenes of impact craters.
Detecting a Mars crater is of significant scientific value to enhance
the understanding of the geomorphological process of the Mars surface,
facilitating various Mars exploration missions for identifying
extraterrestrial life and water on the red planet. We leveraged two resources to create the training data set: (1) the global database of Mars impact craters (greater than or equal to 1 km) containing 384,343 craters, along with their center location and diameter sizes \citep{robbins2012mars}; and (2) the Mars Odyssey thermal emission imaging system--infrared (THEMIS--IR) daytime global mosaic, which provides the imagery of Mars' surface at 100 m resolution \citep{edwards2011mosaicking}.

To create the training images, we randomly selected 10,000 locations and
clipped the same number of image scenes at a size of 256~×~256
\({pixel}^{2}\) (covering an area of 655.36 km\textsuperscript{2}).
According to the location and extent of each image scene, a search
through the Mars crater database was conducted to find the craters
within each scene. Note that a spatial index is built on top of the
database to allow for fast spatial query of the crater data. Image
scenes containing partial craters were discarded. For the craters within
each scene, the count label---number of craters within an image---was
generated for use as the input of our WSOD model. Strong labels (object
class and BBOX) were also created for evaluating the predictive
performance of our proposed model. In this way, we were able to
automatically generate an authentic data set to guide the machine
learning process.

The experiments were conducted on Amazon Elastic Compute Cloud (EC2).
The g3x.xlarge instance with NVIDIA Tesla M60 GPU which has an 8GB
memory was used to run the experiments.

\subsection{Detection Results}\label{detection-results}

\begin{figure}[ht]
  \centering
  \includegraphics [width=0.7\textwidth]{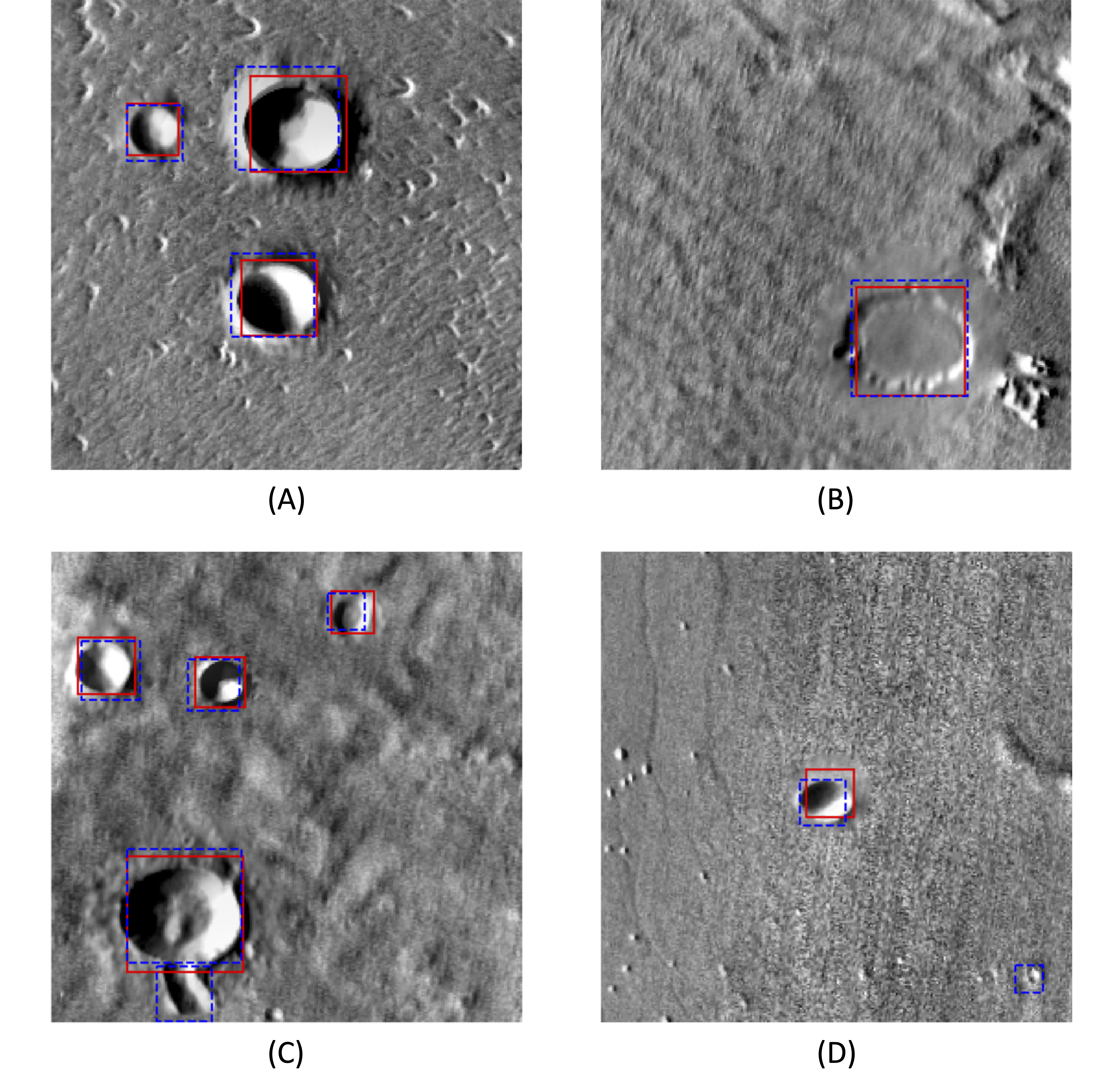}
  \caption{Example detection results. The solid red line is the ground truth bounding box (BBOX), and the dashed blue line is the BBOX predicted by our proposed model.}
  \label{fig:figure4}
\end{figure}

Figure \ref{fig:figure4} demonstrates some sample detection results. Overall, the
network worked pretty well under weakly supervised learning. Using the
Mars crater dataset, our detector achieved 85\% detection accuracy, as
measured by the mean average precision (mAP). The result displayed in
Figure \ref{fig:figure4}(A) shows that craters of different sizes can all be detected
correctly from the image. Even without the provision of the ground truth
BBOX (red), the predicted BBOX is still highly accurate and has
significant overlap with the ground truth. In Figure \ref{fig:figure4}(B), even the
pattern is not obvious, and the difference between the object and the
background is small, but the network still does well with detection.
Furthermore, the network detects craters that are not in the benchmark
dataset. For instance, in Figure \ref{fig:figure4}(C), even though the ``crater'' at the
bottom of the images does not exist in the Mars crater database, our
model is still able to detect it. With this strong detection capability,
scientists can further clarify the relations between this object and
another labeled crater it connects with, for instance, whether they are
both impact craters and belong to the same crater or they should be
labeled as two different craters. Another example is shown in Figure
\ref{fig:figure4}(D). In the original data source, craters with diameters less than 1 km
are not labeled. But there is a vast number of such craters on the Mars
surface; relying on manual labeling of these smaller craters would not
be feasible because the task would be too labor-intensive and could take
years to decades to finish. In Figure \ref{fig:figure4}(D), such a small crater
appearing near the bottom right of the image is successfully detected by
our proposed model. This detection result shows the generalizability of
our approach.

\subsection{Performance Comparison with Cutting-Edge Weakly Supervised
Object Detection
Models}\label{performance-comparison-with-cutting-edge-weakly-supervised-object-detection-models}

In this experiment, our proposed network was compared with cutting-edge
WSOD models, using the same crater detection task to evaluate their
effectiveness. The models compared with our approach included (1) a
weakly supervised deep detection network (WSDDN; \cite{bilen2016weakly}), one of the phenomenal works that addresses the WSOD problem using
CNN; (2) an online instance classifier refinement (OICR; \cite{tang2017multiple}), a popular model that improves WSDDN through multistage
classification refinement; (3) a C-MIDN \citep{gao2019cmidn}, a WSOD model
with cutting-edge performance achieved by utilizing a double WSDDN and a
segmentation map for better proposal selection. Table \ref{tab:table1} demonstrates
both the prediction accuracy and runtime for each model. It can be seen
that our proposed model yielded the overall best performance in terms of
detection accuracy (mAP) among all models. This is due to its smart
spatial principle enhanced strategy for generating proposals on or near
the objects of interest instead of exhausting possible locations using
traditional approaches. Besides, the use of the attention map could
further guide our model to better detect these critical points, yielding
better detection accuracy than both the classic and cutting-edge models.
Regarding model runtime at the prediction phase, as the model is
improved by adding more components to achieve better performance, their
runtime also increases. For instance, OICR is an improvement made on top
of WSDDN by adding an iterative optimization procedure; and C-MIDN
advances OICR by adding a segmentation map for better object
localization and classification. Compared to these cutting-edge models
especially C-MIDN, our proposed model yields the best predictive
accuracy and reasonable runtime.

\begin{table}[ht]
\caption{Comparison of detection performance among different models. A nearly 10\% performance (mAP) increase was achieved with our model, as compared to the cutting-edge C-MIDN model. Our model is also more computationally efficient than C-MIDN. }
\centering
\begin{tabular}{lcc}
\toprule
\textbf{WSOD Models}       & \textbf{mAP (\%)} & \textbf{Run Time (images/sec)} \\
\midrule
WSDDN                     & 62.7              & 2.89 \\
OICR                      & 68.2              & 3.58 \\
C-MIDN                    & 75.8              & 5.41 \\
\textbf{Our proposed model} & \textbf{84.8}     & 4.83 \\
\bottomrule
\end{tabular}

\label{tab:table1}
\end{table}

\subsection{Effect of Optimized Training
Strategies}\label{effect-of-optimized-training-strategies}

\begin{figure}[ht]
  \centering
  \includegraphics [width=1\textwidth]{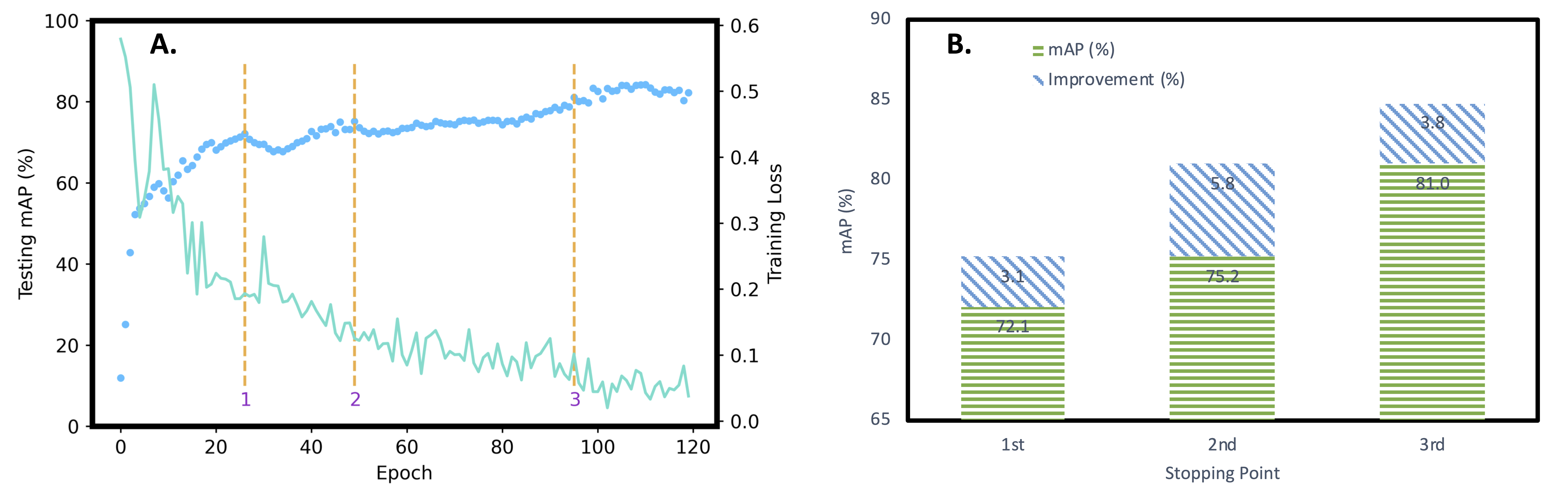}
  \caption{Performance improvement by different training optimization strategies. (A) Training loss (solid line which shows a declining trend) and the predictive accuracy measured by mAP (discrete points showing an increasing trend). (B) Quantitative improvement by proposed optimization strategies: dynamic learning rate (1st), batch normalization (2nd), and reduced candidate proposals (3rd).}
  \label{fig:figure5}
\end{figure}

It is known that for deep neural networks, training strategies are
highly important for achieving a satisfying result. However, there is no
universal guideline about how to train a network, and the training
strategy often varies according to different tasks. In our crater
detection task, instead of tuning multiple hyper-parameters with
multiple models, we adopted another strategy using an experimental
search to train and fine-tune the model. During training, we monitored
the model's performance, and each time the accuracy curve became flat
(meaning that the model was stuck in a local optimum), we saved and
re-trained the model with different hyper-parameters from the point
where the phenomenon occurred. The improvement introduced by our
proposed strategy is illustrated in Figure \ref{fig:figure5}. Dashed vertical lines in
Figure \ref{fig:figure5}A indicate the points we stopped the training and re-trained the
model with different hyper-parameters. At the first stop, we changed the
learning rate. When training a deep neural network, the learning rate is
often the most important factor in deciding a model's performance. A
small learning rate can slow down the learning process and make the
model very difficult to converge. A large value, in comparison, may help
the model to quickly converge but at a suboptimal solution. Hence, we
first set a large value to train the model (0.001), and when the
accuracy curve became flat, we modified it with a smaller value (0.0005)
to increase the search space toward a better solution. A 3.1\% mAP
increase was found by changing the learning rate (Figure \ref{fig:figure5}B).

Next, when the curve became flat again, we integrated batch
normalization, a technique to renormalize the data during training to
further boost performance. To achieve this, we added a batch
normalization layer to renormalize the data before it was used by our
RPN and the object classifier. Adding a normalization layer means the
layers after it need to be re-trained. It can be seen in Figure \ref{fig:figure5}A that
after the second stop, it took a relatively long training period before
the accuracy curve went up again. This strategy yields a 5.8\% increase
in mAP. Finally, at the last stop, we reduced the total number of
proposals. A regular RPN generates proposals (candidate BBOX) with
several ratios. The classifier will then choose the best-fitted proposal
containing the target object. When the number of total proposals
increases, the difficulty of training a classifier increases. Since the
Mars crater dataset only contains nearly round objects, we removed other
ratios to reduce the training efforts of the classifier. As Figure 5B
demonstrates, this modification improved the accuracy by an additional
3.8\%. Figure 5B also quantifies the accuracy increase introduced by
each optimization strategy. Through the experimental search, we were
able to identify the proper training strategies for this task.
Furthermore, additional experiments were conducted, and we found that
these strategies can also be applied in combination at the beginning of
the training to further reduce training time and achieve a satisfying
performance.

\subsection{Effects of Size and Diversity of the Training Data
Set}\label{effects-of-size-and-diversity-of-the-training-data-set}

\begin{figure}[ht]
  \centering
  \includegraphics [width=1\textwidth]{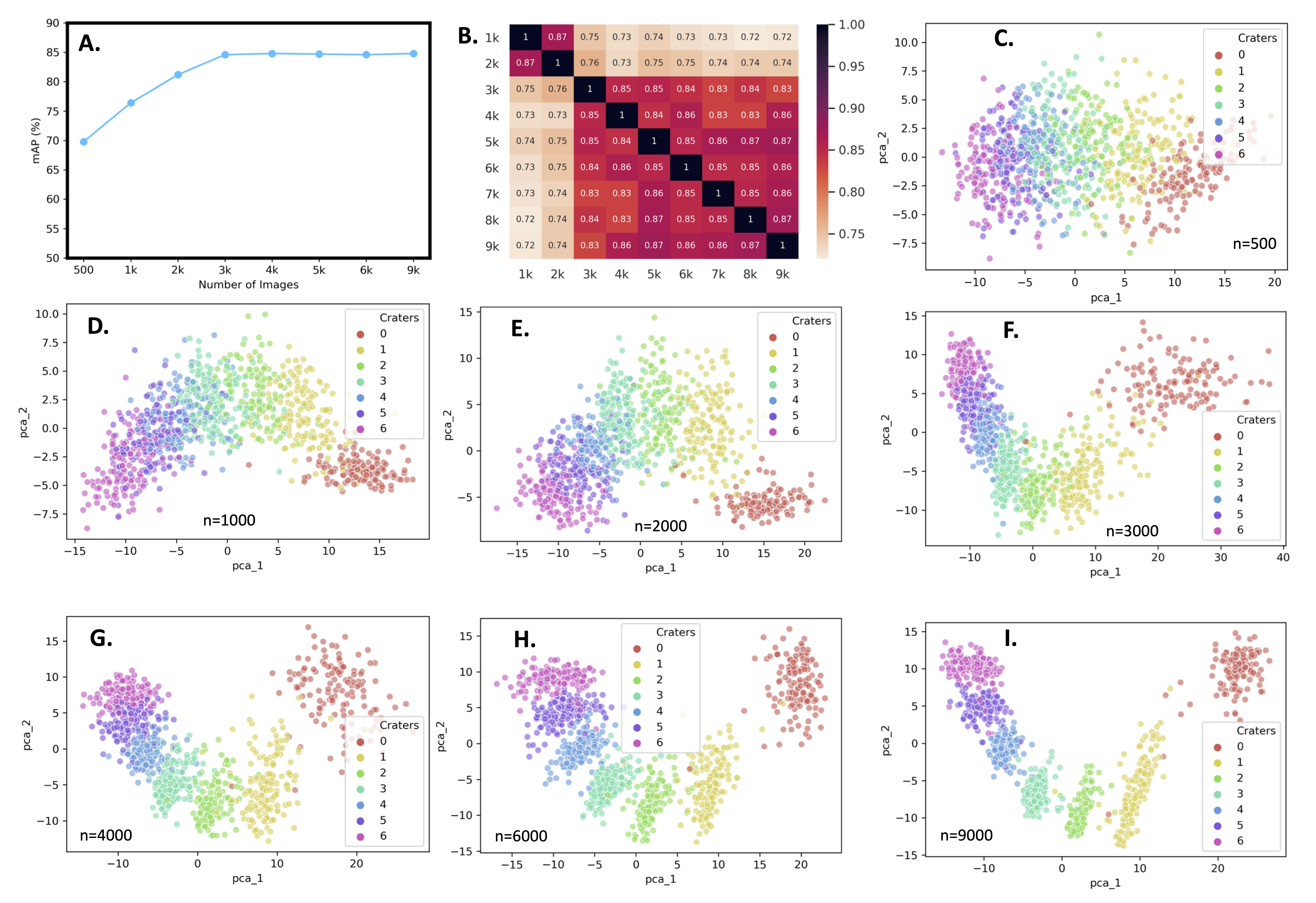}
  \caption{Model performance as an effect of the size of the training data set.}
  \label{fig:figure6}
\end{figure}

In this experiment, we tested the impact of the number of training
images on model performance. While our approach can, in theory, generate
any number of training images, how many would be sufficient, and how
many overabundant? We argue here that the right number of images needed
is the set that can fully represent the data distribution in the
original data source. More images than this number will not contribute
further to the performance but will instead increase training time. To
answer this question, we trained our model with different sets of
training data randomly selected from the original training dataset
containing 9,000 image scenes, and all of these models used the same
extra 1,000 images for testing. The result shown in Figure \ref{fig:figure6} provides
the comparative results. It can be seen that the overall model
performance (prediction accuracy) became stable after the number of
training images reached \emph{n}~=~3,000 and more. To further illustrate
the relationship between the model performance and the number of
training images, we extracted feature maps of the testing images from a
different training set and performed principal component analysis (PCA)
to obtain the prominent features in an abstract space and then projected
the first two components onto a 2D plane (Figure \ref{fig:figure6} C--I). We use colors
to represent the images containing different numbers of craters (0--6)
such that the figures can show not only the overall distribution of the
feature components but also the distribution of images separated by the
number of craters they contain.

These figures demonstrate a better cluster separation and a similar
feature distribution when training sets are 3,000 items and above. In
particular, the better-separated clusters indicate that the model can
better identify different objects and generate discriminating features
that are important for the RPN to achieve object localization. We also
calculated the cosine similarity of image features generated by the
models trained with the above sets. The similarity matrix between
different input sets (\emph{n}~=~1,000--9,000) shown in Figure \ref{fig:figure6}(B)
quantitatively verifies the observation. This result provides guidance
and confidence in selecting the set with 3,000 images to train the model
for equally good performance and better training efficiency.

\subsection{ Model generalizability in object detection of natural
features on Earth and other planets}

To verify the generalizability of our proposed WSOD model, we performed
additional experiments on a natural feature dataset composed of four
terrain categories (crater, hill, volcano and sand dunes). While most of
these features are natural features on Earth, we also added some images
from the Mars (i.e. craters) to diversify the datasets. The training
data contains \textasciitilde120 remote sensing images per terrain
category. Table \ref{tab:table2} provides the prediction accuracy on this new dataset
using different WSOD models. Among all comparable models, ours performs
the best, with the highest AP achieved for crater, volcano and sand
dunes. It also beats other models in the overall mAP. This is attributed
to the novel strategy we developed to enable more accurate object
localization without this information being explicitly provided in the
training data.

\begin{table}[ht]
\caption{Comparison of prediction accuracy in terms of average precision (AP) for individual terrain categories and mean AP (mAP) across all categories.}

\centering
\begin{tabular}{lccccc}
\toprule
\textbf{Models} & \textbf{Crater} & \textbf{Hill} & \textbf{Volcano} & \textbf{Dunes} & \textbf{mAP} \\
\midrule
WSDDN   & 0.68 & 0.59 & 0.66 & 0.45 & 0.60 \\
OICR    & 0.79 & 0.77 & 0.72 & 0.67 & 0.74 \\
C-MIDN  & 0.85 & \textbf{0.79} & 0.79 & 0.71 & 0.79 \\
\textbf{Ours} & \textbf{0.88} & 0.77 & \textbf{0.81} & \textbf{0.73} & \textbf{0.80} \\
\bottomrule
\end{tabular}
\label{tab:table2}
\end{table}

\begin{figure}[ht]
  \centering
  \includegraphics [width=1\textwidth]{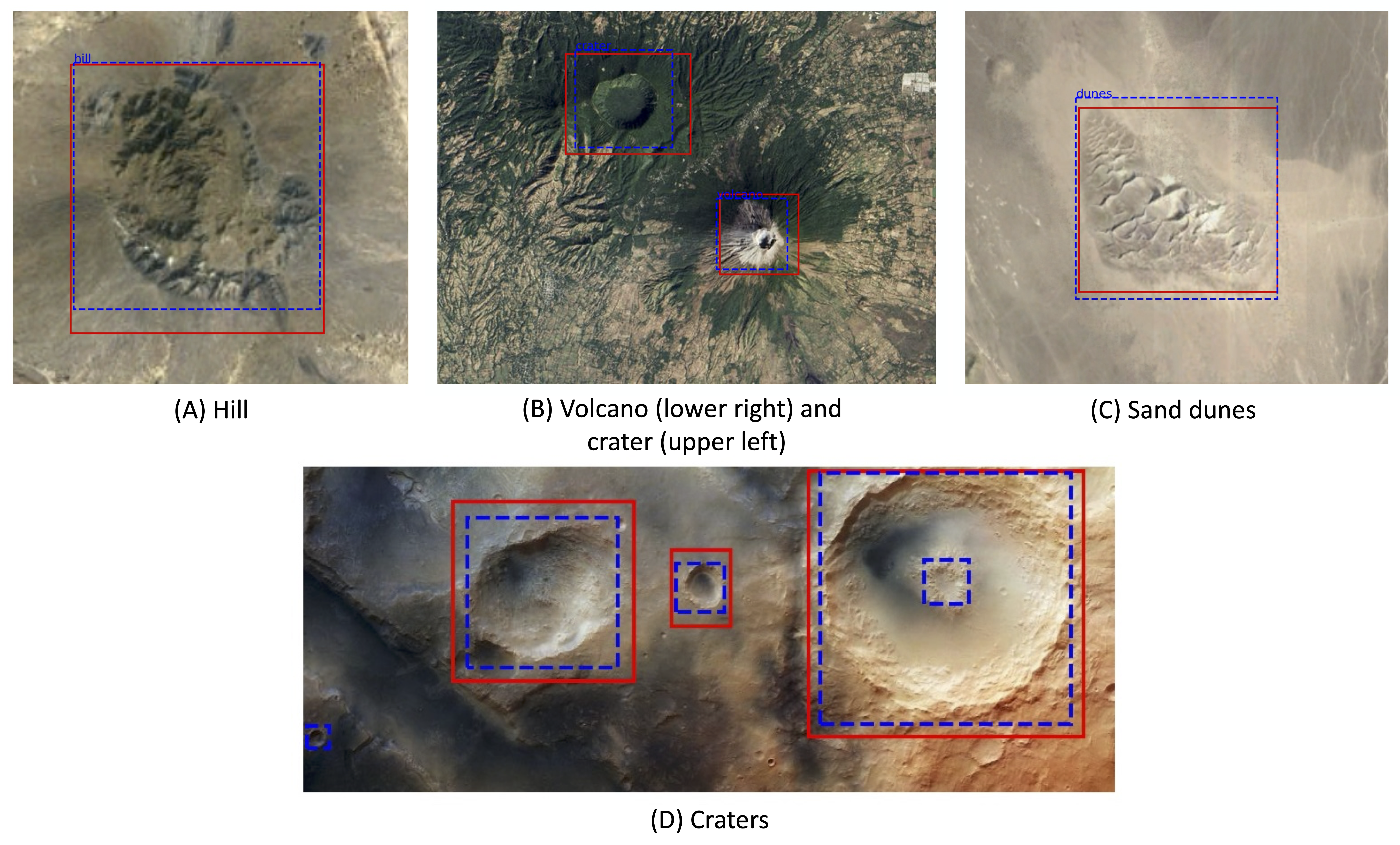}
  \caption{Prediction results for natural terrain features. The solid red line is the ground-truth bounding box (BBOX), and the dashed blue line is the BBOX predicted by our proposed model.}
  \label{fig:figure7}
\end{figure}

Figure \ref{fig:figure7}  demonstrates the prediction results of each feature category
(in dashed blue box) and the ground-truth labels (in solid red box).
Note that the ground-truth labels are only used for results evaluation
and not for model training in the weak supervision context. Almost
perfect object extents are predicted using our proposed approach. The
model is also capable of predicting multiple instances in the same image
scene (Figure \ref{fig:figure7}B). For crater detection, it can be seen that not only simple craters are detected, but nested ones and small craters which do
not exist in the original database can also be detected (Figure \ref{fig:figure7}D).
This result clearly demonstrates the effectiveness of the proposed
approach in detecting natural features.

\section{Summary and Outlook}\label{summary-and-outlook}

This paper reports a brand-new attempt to develop a WSOD framework using
in-depth integration of cutting-edge deep learning models and spatial
theory and principles. It is known that natural feature detection has suffered tremendous challenges due to the lack of proper training data,
the various forms of objects of the same class, and their vague
boundaries, as compared to manmade features. This paper tackles the
problem with an innovative solution that converts the 2D object
detection problem into a 1D temporal classification to utilize the
latter's advanced search optimization technique, which allows for the
use of weak labels for the detection task. What enables such a
conversion is spatial autocorrelation as stated in the Tobler's first law of geography. Utilizing the scan order
along x or y direction, the spatial continuity along one direction is
retained, and its continuity along the perpendicular direction will
still be captured by the long- and short-term memory of the RNN (i.e.,
LSTM). Experiments were conducted to prove that our proposed model
achieved state-of-the-art performance. 

In the future, we will further
refine our model and leverage its capabilities to enrich the existing
Mars crater databases. As the most comprehensive database containing
over 380,000 impact craters on Mars, the one developed by Robbins and
Hynek \cite{robbins2012mars} was the result of multiyear research containing tremendous manual work and labeling efforts. However, a large number of small
craters with diameters less than 1 km have not yet been included in the
database. Our work in AI and especially GeoAI will play a key role in
automating the detection of Mars craters, as well as many other natural
features on Earth's surface, especially with the lack of proper training
data. This research provides direct GeoAI support to science for improved exploration of Earth and space \citep{li2025artificial, li2024geoai}.

\section*{Acknowledgment}
This work is in part supported by the National Science Foundation under grants BCS- 1853864, BCS-1455349, OIA-2033521, OIA-1936677, and OIA- 1937908.

\bibliographystyle{abbrv}  
\bibliography{bibtex/bib/main}

\end{document}